\documentclass{article}

\usepackage[preprint]{neurips_2025_modified}

\usepackage[utf8]{inputenc} 
\usepackage[T1]{fontenc} 
\usepackage{hyperref} 
\usepackage{url} 
\usepackage{booktabs} 
\usepackage{amsfonts} 
\usepackage{nicefrac} 
\usepackage{microtype} 
\usepackage{xcolor} 

\usepackage{algorithm}
\usepackage{algpseudocode}
\usepackage{amsmath}
\usepackage{tabularx}
\usepackage{array}
\usepackage{graphicx}
\usepackage{colortbl}
\usepackage{enumitem}
\usepackage{placeins}
\usepackage{tikz}
\usepackage{amssymb}
\usetikzlibrary{arrows.meta, positioning, shapes.geometric, fit, shadows, shapes.callouts, decorations.pathmorphing}
\definecolor{dodgerblue}{RGB}{30,144,255}

\newcommand{\pdesc}{\mathcal{P}_{\textrm{desc}}}
\newcommand{\peval}{\mathcal{P}_{\textrm{eval}}}
\newcommand{\psol}{\mathcal{P}_{\textrm{sol}}}
\newcommand{\cotmethod}{\textsc{CoT}}
\newcommand{\dcmethod}{\textsc{CreativeDC}}
\newcommand{\basemethod}{\textsc{Base}}
\newcolumntype{C}[1]{>{\centering\arraybackslash}m{#1}}
\providecommand{\Description}[1]{}

\title{Enhancing Diversity of LLM-Generated\\ Educational Tasks}

\author{
Manh Hung Nguyen\\
MPI-SWS, Germany\\
\text{manguyen@mpi-sws.org} \\
\And Sebastian Tschiatschek \\
University of Vienna, Austria \\
\text{sebastian.tschiatschek@univie.ac.at} \\
\And Adish Singla \\
MPI-SWS, Germany \\
\text{adishs@mpi-sws.org} \\
}

\begin{document}
    \maketitle
    
\begin{abstract}

\looseness-1Large language models (LLMs) have shown the potential for generating educational content at scale, assisting educators in creating practice tasks or synthesizing data for training educational models. However, LLMs suffer from the ``Artificial Hivemind'' effect, where they produce homogeneous content. This homogeneity limits the diversity of LLM-generated tasks, a crucial factor in these educational settings. In this paper, we investigate how to increase the diversity of generated tasks while keeping their utility high. Inspired by the divergent--convergent thinking stages in creativity literature, we propose a prompting framework with two reasoning stages: (1) exploring the creative space, and (2) satisfying the input requirements. We evaluate {\dcmethod}, a method instantiated from this framework in the domain of Python programming, using both automated metrics and expert evaluation. Results show that {\dcmethod} produces significantly more distinct high-utility tasks (about $1.6\times$) than baselines. Our work offers an effective approach for generating and evaluating more diverse tasks at scale.
\end{abstract}


\section{Introduction}
\label{sec:introduction}
\begin{figure*}[t]
    \Description{A flowchart illustrating the CreativeDC method. The context goes into a divergent thinking box producing 15 diverse ideas, one of which goes into a convergent thinking box that refines it into an output programming task.}
    \centering
    \begin{tikzpicture}[
    node distance=0.3cm,
    box/.style={
        rectangle,
        draw=black,
        thick,
        rounded corners=3pt,
        align=center,
        font=\sffamily
    },
    promptcontent/.style={
        box,
        minimum width=3.0cm,
        minimum height=1.5cm,
        text width=3.0cm,
        font=\sffamily\scriptsize
    },
    titlebox/.style={
        box,
        minimum width=5cm,
        minimum height=0.8cm,
        font=\sffamily\bfseries
    },
    contentbox/.style={
        box,
        minimum width=5cm,
        minimum height=6cm,
        text width=5.5cm,
        align=justify,
        font=\sffamily\scriptsize
    },
    outputbox/.style={
        box,
        minimum width=9.2cm,
        minimum height=3.8cm,
        text width=9.2cm,
        font=\sffamily\scriptsize,
        fill=white,
        drop shadow
    },
    ideabox/.style={
        box,
        minimum width=9.2cm,
        minimum height=2.2cm,
        text width=9.2cm,
        align=left,
        font=\sffamily\scriptsize,
        fill=white,
        drop shadow
    },
    convergentbox/.style={
        box,
        minimum width=9.2cm,
        minimum height=3.0cm,
        text width=9.2cm,
        align=left,
        font=\sffamily\scriptsize,
        fill=white,
        drop shadow
    },
    dctitle/.style={titlebox, fill=dodgerblue!50, text=black, minimum width=2.0cm, font=\small},
    dccontent/.style={contentbox, minimum width=3.4cm, minimum height=4cm, text width=3.0cm},
    arrow/.style={
        -Stealth,
        thick,
        shorten >=2pt,
        shorten <=2pt
    },
    icon/.style={
        circle,
        fill=#1,
        minimum size=0.5cm,
        inner sep=0pt
    },
    iconlabel/.style={
        font=\sffamily\scriptsize,
        align=center
    },
    problemtitle/.style={
        font=\sffamily\bfseries\scriptsize
    }
]

\definecolor{dodgerblue}{RGB}{30,144,255}

\node[dctitle] (dctitle) {\dcmethod};
\node[dccontent, below=0cm of dctitle] (dccontent) {

    Apply the following thinking process.\\[0.3em]

    \textbf{Divergent thinking stage:} Think about different elements, objects, scenarios, or situations that are relevant to the given theme. Push for unusual, surprising, unconventional, and diverse ideas.\\[0.3em]

    \textbf{Convergent thinking stage:} From your brainstormed ideas, select one and connect it with the required programming concepts to create a creative programming task. If it does not work, try another idea.\\[0.3em]

    Given a theme of Cooking, create a Python programming task that requires only Loops to solve.
};

\node[fit=(dctitle)(dccontent), inner sep=0pt] (dcblock) {};

\node[promptcontent, above=0.5cm of dcblock] (promptcontent) {
    \textbf{Context} \\[0.3em]
    
    Theme: Cooking\\
    Concept: Loops
};

\node[ideabox, right=0.6cm of dcblock, yshift=5.0cm, anchor=west, align=justify] (divergentoutput) {
    \textbf{(1) Divergent Thinking:}\\[0.5em]
    1. A sentient kitchen appliance that whispers recipes only at midnight.\\
    2. A stew that must be stirred in prime-numbered intervals to avoid bitterness.\\
    3. A curry that changes flavor based on the cook's emotions, recorded via biometrics.\\
    4. A recipe book that erases itself after each use unless you sing to it.\\
    ...\\
    15. A kitchen garden where vegetables grow only when spoken to in rhyming couplets.
};

\node[convergentbox, below=0.5cm of divergentoutput.south west, anchor=north west, align=justify] (convergentoutput) {
    \textbf{(2) Convergent Thinking:}\\[0.5em]
    Selected idea: A stew that must be stirred in prime-numbered intervals to avoid bitterness. Conceptual link: The cook must simulate stirring a stew at every minute from 1 to N, but only stirs when the minute number is a prime. The bitterness level increases by 1 for every non-prime minute skipped. The goal is to compute the total bitterness accumulated over N minutes if the cook only stirs on prime minutes. This naturally maps to using loops to check each minute, determine if it's prime via loop-based primality test, and accumulate bitterness. The solution requires only loops (no recursion, conditionals beyond basic if, lists, or external libraries).
};

\node[outputbox, below=0.5cm of convergentoutput.south west, anchor=north west] (dcoutput) {};

\node[problemtitle, below=0.0cm of dcoutput.north] (dc_problem) {Generated Task};
\node[iconlabel, below=0.8cm of dcoutput.north, xshift=3.95cm] (dc_test_lbl) {\textbf{Tests}};
\node[below=-0.08cm of dc_test_lbl] (dc_test) {\includegraphics[width=0.5cm]{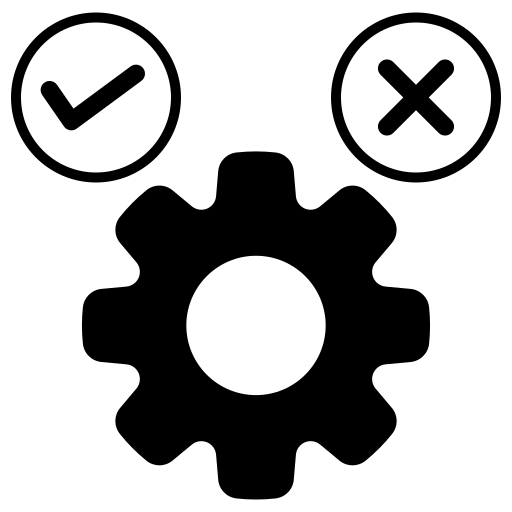}};
\node[iconlabel, below=-0.05cm of dc_test] (dc_sol_lbl) {\textbf{Solution}};
\node[below=-0.08cm of dc_sol_lbl] (dc_sol) {\includegraphics[width=0.4cm]{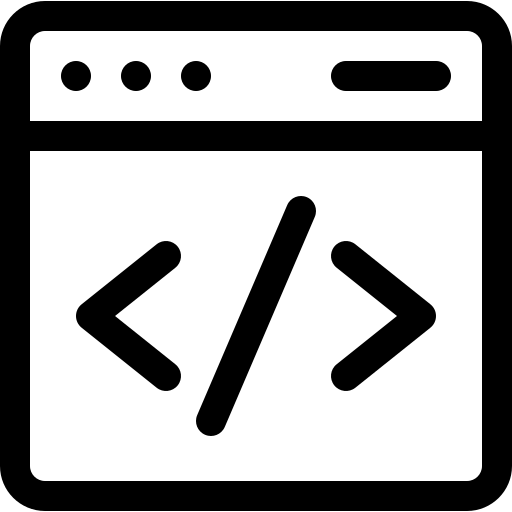}};
\node[font=\sffamily\scriptsize, text width=8.0cm, align=justify, below=0.4cm of dcoutput.north, xshift=-0.6cm] (dc_text) {\hspace{0.1cm}\raisebox{-0.1cm}{\includegraphics[width=0.4cm]{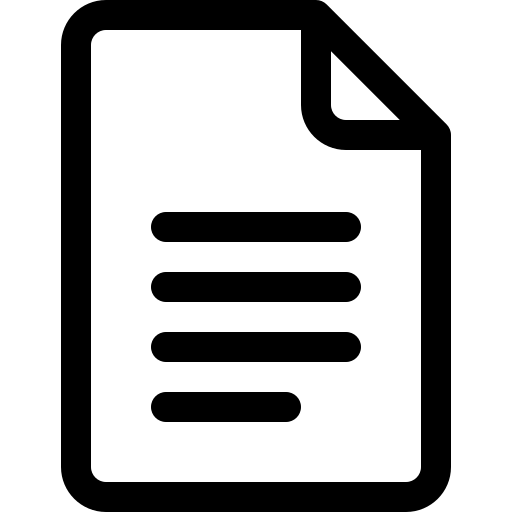}}\hspace{0.15cm}\textbf{Description:} You are managing a magical stew that must be stirred exactly on prime-numbered minutes to prevent it from becoming bitter. The cooking process lasts for N minutes, starting from minute 1 to minute N. Each time the stew is not stirred on a non-prime minute, its bitterness level increases by 1. However, on prime-numbered minutes, you stir the pot, and no bitterness is added. Your task is to write a function called `calculate\_bitterness` that takes one argument: the total cooking time in minutes (an integer N). The function should return the total bitterness level of the stew after N minutes. Assume N is a positive integer greater than or equal to 1. The function will be tested with various values of N to ensure correctness.};

\node[draw=black, thick, rounded corners=3pt, fit=(divergentoutput) (convergentoutput) (dcoutput), inner sep=0.15cm] (containerbox) {};

\draw[arrow] (promptcontent.south) -- (dcblock.north);

\draw[arrow] (dcblock.east) -- (containerbox.west |- dcblock.east);

\draw[arrow] (divergentoutput.south) -- (convergentoutput.north);
\draw[arrow] (convergentoutput.south) -- (dcoutput.north);

\end{tikzpicture}
    \caption{\textbf{Overview of {\dcmethod} for programming task generation}. The input is a context containing a theme and a programming concept. Our method instructs an LLM to apply a two-stage reasoning process: \textit{(1) Divergent thinking stage} and \textit{(2) Convergent thinking stage}. The generated task consists of a description, a test suite, and a solution.}
\label{fig:illustrative_example}
\end{figure*}
\looseness-1Large Language Models (LLMs) have demonstrated remarkable capabilities, leading to their rapid adoption across important domains such as education~\cite{DBLP:conf/aaai/ElkinsKCS24,DBLP:conf/edm/0006S0L24,DBLP:conf/lak/LiCB25,DBLP:conf/aied/BulathwelaMY23,DBLP:conf/emnlp/0001LB21,DBLP:conf/lats/LiGLX24,DBLP:conf/aied/JiaoSCZS23,DBLP:conf/aied/YuKL25,DBLP:conf/edm/KarbasiHSP25,DBLP:conf/ectel/ScholzNSN25,DBLP:conf/aied/DemirtasZFC25,DBLP:conf/aied/ScarlatosLLBL25,DBLP:conf/edm/NguyenTS24,DBLP:conf/sigcse/SolanoK0VR26}. In particular, prior work has used LLMs to generate programming tasks~\cite{DBLP:conf/aied/NguyenPGTS25,DBLP:conf/icer/LogachevaHPS024,DBLP:conf/icer/SarsaDH022,DBLP:conf/lak/HassanyBSNAAH25}, enabling active learning strategies such as the ``doer effect'' through task generation at scale~\cite{DBLP:journals/corr/abs-2402-01580,campenhout2025scaling}. Moreover, LLMs can also be used to generate synthetic data, for example, for training tutor models~\cite{DBLP:conf/aied/ScarlatosLLBL25,DBLP:conf/sigcse/SolanoK0VR26,DBLP:journals/corr/abs-2406-20094}. In such settings, maintaining task diversity is crucial to adequately cover learning objectives and different difficulty levels. Yet LLMs have been shown to exhibit an ``Artificial~Hivemind'' effect~\cite{hivemind}, where they tend to produce similar and repetitive outputs within a single model and across different models, resulting in limited output diversity.

\looseness-1Existing work has proposed various methods to address the limited diversity and homogeneity of LLM outputs. These include fine-tuning through preference optimization with creativity signals~\cite{ismayilzada-etal-2025-creative} and using multi-agent discussion or debate frameworks~\cite{lu2024discussion}. However, these approaches are typically designed and evaluated for benchmarks such as creative writing or general instruction following, rather than the structured requirements of educational task generation. Another line of work has incorporated thematic topics and learning concepts as input to LLMs to generate a wide range of tasks, e.g., in programming education~\cite{DBLP:conf/aied/NguyenPGTS25,DBLP:conf/icer/LogachevaHPS024,DBLP:conf/icer/SarsaDH022}. However, these efforts do not primarily focus on optimizing for task diversity. This gap motivates our central question: \textit{How can we generate a diverse set of high-utility practice tasks}? 

In this work, we tackle this question in the context of programming task generation. Drawing inspiration from the creativity literature and the divergent-convergent thinking stages~\cite{wallas1926art,10.7551/mitpress/7722.001.0001,guilford1956structure}, we propose a two-stage prompting framework and instantiate {\dcmethod}, a method for generating diverse contextualized programming tasks. Given a theme and a programming concept as context, {\dcmethod} instructs the LLM to first go through a \emph{divergent thinking stage}, where it explores the creative space of ideas with respect to the input theme. Then, in the following \emph{convergent thinking stage}, the model refines promising ideas into a valid programming task using the given programming concept. This decoupling encourages the LLM to explore a broader ideation space before committing to a final task. Our main contributions are:
\begin{enumerate}[leftmargin=*]
    \item We propose a novel prompting framework with divergent-convergent thinking stages and develop {\dcmethod}, a method for generating diverse programming tasks. 
    \item We define and propose approaches to evaluate {effective diversity}, which measures the number of distinct tasks among high-utility tasks.
    \item We show through automated and expert evaluations that {\dcmethod} achieves significantly higher effective diversity than baseline methods. 
\end{enumerate}

Figure 1 shows an overview of our method. We release our code and data publicly to support future research at: \href{https://github.com/machine-teaching-group/edm2026-creativedc}{https://github.com/machine-teaching-group/edm2026-creativedc}.


\section{Related Work}
\label{sec:relatedwork}

\textbf{LLMs for Task Generation.}
LLMs have been used to generate educational tasks across domains, including topic-specific questions, mathematics word problems, and programming exercises~\cite{DBLP:conf/lak/LiCB25,DBLP:conf/aied/BulathwelaMY23,DBLP:conf/emnlp/0001LB21,DBLP:conf/lats/LiGLX24,DBLP:conf/edm/0006S0L24,DBLP:conf/aied/JiaoSCZS23,DBLP:conf/aied/YuKL25,DBLP:conf/edm/KarbasiHSP25}. In programming education, recent work has generated multiple-choice items and contextualized tasks~\cite{DBLP:conf/lak/HassanyBSNAAH25,DBLP:conf/aied/NguyenPGTS25,DBLP:conf/icer/LogachevaHPS024,DBLP:conf/icer/SarsaDH022}. However, these works primarily optimize task correctness or difficulty alignment, whereas we also consider the diversity of generated tasks.

\looseness-1\textbf{Diversity Collapse of LLMs.} LLMs exhibit an ``Artificial Hivemind'' effect, characterized by high intra-model and inter-model homogeneity~\cite{hivemind}. Post-training alignment further amplifies this convergence~\cite{zhang2025noveltybench} and reduces LLM creativity and output novelty~\cite{DBLP:conf/iclr/LuSHM0HEJCD025}. Moreover, LLMs have been shown to lack novelty~\cite{ismayilzada-etal-2025-creative,DBLP:conf/iclr/LuSHM0HEJCD025} and suffer from ``functional fixedness'', a bias that limits unconventional thinking~\cite{DBLP:conf/naacl/TianRQ0MP00B24}. These findings motivate methods that explicitly scaffold exploration rather than direct prompting alone.

\looseness-1\textbf{Improving Diversity of {LLM} Outputs.} Methods have been proposed to promote diversity in LLM outputs, including fine-tuning with creativity signals~\cite{ismayilzada-etal-2025-creative}, multi-agent frameworks~\cite{lu2024discussion,DBLP:journals/corr/abs-2510-07064}, and persona simulation for diverse perspectives~\cite{DBLP:journals/corr/abs-2406-20094}. However, these methods are primarily designed for open-ended generation settings such as creative writing. Our work instead targets educational task generation, where outputs must balance diversity with utility.

\section{Problem Setup}
\label{sec:problemsetup}

We formalize the problem of generating diverse tasks, discuss the metrics for evaluation, and state our objective.

\textbf{Preliminaries.} We denote a programming task as a tuple $\mathcal{P} = (\pdesc, \peval, \psol)$, where ~$\pdesc$ is a natural-language task description, ~$\peval$ is an evaluation guide or final answer for checking the correctness of any answer, and ~$\psol$ is a reference solution. Given a context \(\mathcal{C}\), which specifies the requirements for task generation, and an integer \(K\), we seek to generate a set of tasks that satisfy \(\mathcal{C}\). We denote the generated set of \(K\) tasks as \(\mathcal{S}(\mathcal{C}) = \{\mathcal{P}_1, \mathcal{P}_2, \ldots, \mathcal{P}_K\}\). For example, if \(\mathcal{C}\) corresponds to a learning concept, we aim to generate \(K\) tasks that help students practice that concept.

\looseness-1\textbf{Metrics.} We evaluate a task set $\mathcal{S}(\mathcal{C})$ along the following metrics. \emph{Diversity} captures the degree of task variation and is defined as the number of semantically distinct tasks in $\mathcal{S}(\mathcal{C})$. \emph{Utility} captures whether a task $\mathcal{P} \in \mathcal{S}(\mathcal{C})$ is valid, comprehensible, and relevant to the given context. We assign $\text{Utility}(\mathcal{P}) = 1$ if it meets these criteria and $0$ otherwise. We define the subset of high-utility tasks as $\mathcal{S}_{\text{util}=1}(\mathcal{C}) = \{\mathcal{P} \in \mathcal{S}(\mathcal{C}) \mid \text{Utility}(\mathcal{P})=1 \}$. We define \emph{Effective Diversity} as the diversity of this high-utility subset. Intuitively, it measures the number of distinct high-utility tasks. We detail the measurement of these metrics in Section~\ref{sec:evaluation_setup}.

\textbf{Objective.} Our goal is to design a method that, given a context $\mathcal{C}$, produces a set of $K$ tasks $\mathcal{S}(\mathcal{C})$ with high \emph{effective diversity}. This objective is challenging because encouraging diverse outputs can conflict with the utility constraints~\cite{padmakumar2026measuring}.

In this work, we instantiate this setup for the generation of programming tasks given a context, following prior work on contextualized programming tasks~\cite{DBLP:conf/aied/NguyenPGTS25,DBLP:conf/icer/LogachevaHPS024,DBLP:conf/icer/SarsaDH022}. Figure~\ref{fig:illustrative_example} illustrates an example context and a generated task.

\renewcommand{\arraystretch}{1.2}
\begin{table*}[t]
\caption{Prompts for all methods. \textcolor{gray}{\{placeholder\}} is replaced with the actual theme and concept for each context.}

\label{table:prompts_all}
\small
\begin{tabularx}{\linewidth}{X}
\hline
\rowcolor{gray!15} \multicolumn{1}{c}{\basemethod} \\
\# Task Instruction \\

Given a theme of \textcolor{gray}{\{theme\}}, create a Python programming problem that requires only \textcolor{gray}{\{concept\}} to solve. The problem should include a problem description, a test suite, and a solution program. Below are the requirements for the problem:
\begin{itemize}[leftmargin=*, nosep]
\item The problem must be clearly relevant to the given theme of \textcolor{gray}{\{theme\}} and the theme is explicitly used throughout. It requires only \textcolor{gray}{\{concept\}} to solve the problem.
\item The problem description must be sensible and sound natural. It must provide comprehensive information required to solve the problem and pass the test suite (e.g., how the program will be tested, function signatures). Do not use type hints. Do not mention the required programming concepts in the problem description.
\item The test suite must consist of at least 5 comprehensive test cases written in the Pytest framework format. The testsuite must be correct and cover both base and corner cases. If the test suite involves handling files and I/O, the related files should be created using `setup\_module()' and removed using `teardown\_module()' functions in the Pytest framework. Everything from the solution program will be imported manually; do not import anything else except `pytest' and `os'. Do not use multiple assert statements in a single test case.
\item The solution program must use only \textcolor{gray}{\{concept\}}. Do not include any comments, usage examples, or tests in the solution program. The solution program must pass the test suite.
\end{itemize}

\# Output Format \\

Output a JSON object with the following keys: \texttt{`description'}, \texttt{`test\_suite'}, and \texttt{`solution'}. \\
\hline
\rowcolor{red!15} \multicolumn{1}{c}{\cotmethod} \\
\emph{Think step by step to generate a problem.} \\
\# Task Instruction (same as in \basemethod) \\

\# Output Format \\

Output a JSON object with the following keys: \texttt{`chain\_of\_thought'}, \texttt{`description'}, \texttt{`test\_suite'}, and \texttt{`solution'}. \\
\hline
\rowcolor{dodgerblue!30} \multicolumn{1}{c}{\dcmethod} \\
\textit{Apply the following thinking process to generate a problem.} \\
\textit{Divergent thinking phase:} Think about only the given theme and list down wildly different and underexplored elements, objects, scenarios, or situations that are relevant. Ignore the required programming concepts in this phase. Push for unusual, surprising, unconventional, and diverse ideas. Explore the creative space related to the theme as much as possible. \\
\textit{Convergent thinking phase:} From your brainstormed ideas, select one and connect it with the required programming concepts to create a creative programming problem. Make sure the problem does not require any other programming concepts other than the given programming concepts. If it does not work, feel free to go back to select another idea and try again. \\
\# Task Instruction (same as in \basemethod) \\

\# Output Format \\

Output a JSON object with the following keys: \texttt{`divergent\_thinking'}, \texttt{`convergent\_thinking'}, \texttt{`description'}, \texttt{`test\_suite'}, and \texttt{`solution'}. \\
\hline
\end{tabularx}
\end{table*}

\section{Our Method: {CreativeDC}}
\label{sec:methodology}
We draw inspiration from creativity literature and staged models~\cite{wallas1926art,10.7551/mitpress/7722.001.0001,guilford1956structure} to propose a two-stage prompting framework that structures the model's reasoning into \emph{divergent thinking} and \emph{convergent thinking} stages. Divergent thinking aims to generate many diverse ideas without judgment, while convergent thinking selects and refines the best ideas into a concrete solution. We instantiate this framework as {\dcmethod} (\textbf{Creative} \textbf{D}ivergent-\textbf{C}onvergent) (cf. Figure~\ref{fig:illustrative_example}).

\textbf{Stage 1: Divergent Thinking.} In this stage, {\dcmethod} instructs the LLM to explore the thematic space of context $\mathcal{C}$ without imposing any constraints from the target programming concept in $\mathcal{C}$. By focusing only on the theme, the model is encouraged to brainstorm varied and underexplored elements, objects, scenarios, or situations. Pushing for unusual, surprising, and unconventional ideas helps the model traverse a broader semantic space.  The goal of this stage is to avoid the premature convergence often seen in direct prompting, where the model settles too quickly on ideas that satisfy the technical requirements but offer limited variety.

\textbf{Stage 2: Convergent Thinking.} In the second stage, {\dcmethod} instructs the model to select one promising idea from the brainstormed list and refine it into a valid programming task $\mathcal{P}$. The output must satisfy all requirements of $\mathcal{C}$ and, crucially, primarily test the specified programming concept. If the selected idea cannot be developed into a valid task, the model is encouraged to return to the brainstormed list and try another idea. In this way, the convergent stage uses the broader idea pool produced by Stage 1 while enforcing the constraints required for utility. 

After going through the two stages, the model outputs the final task $\mathcal{P} = (\pdesc, \peval, \psol)$. Figure~\ref{fig:illustrative_example} illustrates the two stages of our method with an example generated task.


\section{Evaluation Setup}
\label{sec:evaluation_setup}

In this section, we present an evaluation setup designed to answer the following questions: \textbf{(i)} How does effective diversity scale with the number of generated tasks? \textbf{(ii)} How do expert evaluators perceive the generated tasks?

\textbf{Task Contexts.} We use a set of 20 task contexts, each of which is a pair of a theme and a target programming concept following prior work on contextualized tasks~\cite{DBLP:conf/aied/NguyenPGTS25,DBLP:conf/icer/LogachevaHPS024,DBLP:conf/icer/SarsaDH022}. We use 4 themes (``Cooking'', ``Science Fiction'', ``Superheroes'', and ``Board Games'') and 5 concepts (``Variables'', ``Selection Statements'', ``Loops'', ``Lists'', and ``Strings''). For each context, we generate $K$ tasks per method. We report the results for $K \in \{10, 20, 30, 40, 50\}$ in Section~\ref{sec:evaluation_results}.

\looseness-1\textbf{Methods.} We compare {\dcmethod} (described in Section~\ref{sec:methodology}) against two baselines: {\basemethod} and {\cotmethod}. {\basemethod} directly prompts the model to generate a task from the given theme and concept. {\cotmethod} prepends a ``Think step by step'' instruction to the {\basemethod} prompt to elicit intermediate reasoning~\cite{DBLP:conf/nips/Wei0SBIXCLZ22}. Table~\ref{table:prompts_all} shows the prompts for all methods. For all methods, we use Qwen3-235B-A22B~\cite{DBLP:journals/corr/abs-2505-09388}, a strong open-weight model, enabling reproducible generation. We set the temperature to $1.0$.

\begin{figure}[t]
    \Description{A horizontal diagram showing the user flow through three expert evaluation sessions.}
    \centering
    \includegraphics[width=0.82\linewidth, trim=0cm 6.5cm 0cm 6.85cm, clip]{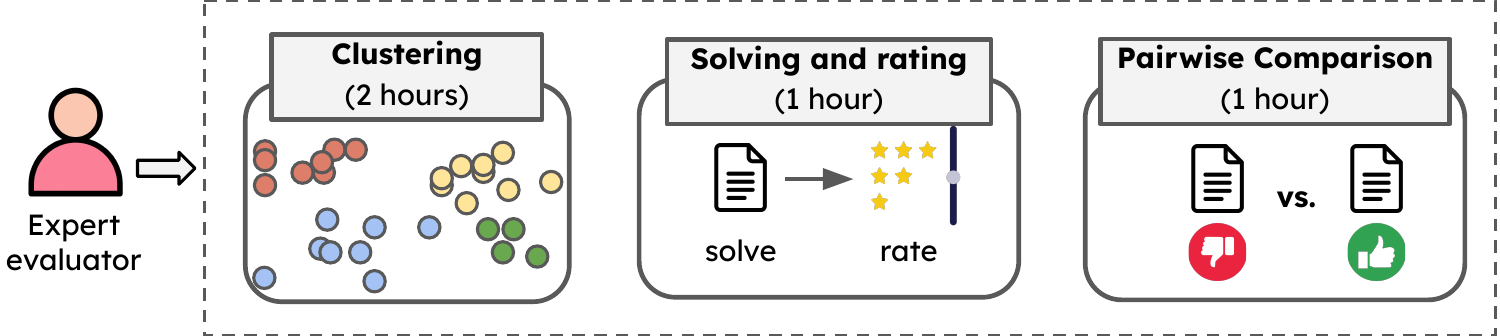}
    \caption{\textbf{Overview of Expert Evaluation.} }
    \label{fig:study_overview}
\end{figure}

\looseness-1\textbf{Expert-based Evaluation Overview.} We use a mix of automated and expert-based evaluation as further discussed in Section~\ref{sec:measuring_effective_diversity} and Section~\ref{sec:measuring_other_task_aspects}. We present an overview of our expert evaluation for assessing diversity and multiple task aspects in Figure~\ref{fig:study_overview}. More concretely, we recruited three experts with experience in programming education and tutoring in programming-related courses. All three are non-native English speakers (one female, two males; mean age 28.7 years), and none are authors of this paper. Participation was voluntary, and no personally identifiable data was collected. Each expert evaluator was assigned the same programming concept (``Variables'') and a randomly sampled theme. We use small subsets of tasks generated during the automated evaluation to reduce the workload. We exclude {\basemethod} from expert evaluation due to similar performance to {\cotmethod} in automated evaluations (cf. Section~\ref{sec:evaluation_results}). To maintain objectivity, the source of each task remained undisclosed throughout the study.

\subsection{Measuring Effective Diversity}
\label{sec:measuring_effective_diversity}
We measure the effective diversity of a task set $\mathcal{S}$ (defined in Section~\ref{sec:problemsetup}), which requires evaluating both task diversity and task utility. Below, we present two approaches we used to evaluate the set diversity: automated and expert-based. 

\begin{itemize}[leftmargin=*]
    \item \textbf{Automated Vendi Score.} We use the Vendi Score~\cite{DBLP:journals/tmlr/FriedmanD23} to measure task diversity automatically. Given a set of tasks, it ranges from $1$ (all tasks identical) to $|\mathcal{S}|$ (all tasks dissimilar), quantifying the number of distinct tasks in $\mathcal{S}$. To begin, each task $\mathcal{P}$ is represented by an embedding obtained from a model (Qwen3-Embedding-0.6B~\cite{DBLP:journals/corr/abs-2506-05176}). Given the task embeddings, we construct a cosine similarity matrix and compute the Vendi Score as the exponential of the Shannon entropy of its eigenvalues.  

    \item \looseness-1\textbf{Expert-based Clustering.} We involve expert evaluators to assess task diversity through clustering the generated tasks. Given a set of tasks, an evaluator either assigns each task to a cluster based on its semantic similarity to other tasks in the set or creates a new cluster for the task. The number of final clusters provides an expert-annotated measure of task diversity, corresponding to the number of distinct tasks as judged by experts. 
    
\end{itemize}

Next, we describe how we measure effective diversity which is defined as the diversity (number of distinct tasks) of the high-utility subset $\mathcal{S}_{\text{util}=1}(\mathcal{C})$ (cf. Section~\ref{sec:problemsetup}). To this end, we need to evaluate the utility of generated tasks. For scalability, we use an LLM-as-a-judge approach, which has been used in prior work including task utility evaluation~\cite{hivemind,DBLP:conf/aied/NguyenPGTS25,padmakumar2026measuring}. Given a task $\mathcal{P}$, the judge first generates a solution, which is executed against $\peval$ to verify \textit{Task Validity}. The judge then assesses \textit{Context Relevance} (whether the task is relevant to the given context) and \textit{Comprehensibility} (whether sufficient information is provided). $\text{Utility}(\mathcal{P}) = 1$ if the task $\mathcal{P}$ satisfies all three criteria. We use Gemini 2.5 Flash-Lite~\cite{DBLP:journals/corr/abs-2507-06261} as the judge, chosen for its cost-effectiveness. Using a model from a different family than the generator also avoids self-preference bias.

\subsection{Measuring Other Task Aspects}
\label{sec:measuring_other_task_aspects}
Beyond diversity metrics, we asked experts to perform additional evaluations on small subsets of generated tasks. More specifically, each evaluator solved and rated $6$ tasks ($3$ from {\cotmethod} and $3$ from {\dcmethod}) via a web interface (providing a task description, code editor, and console). After completing a task or reaching a 10-minute limit, evaluators rated \textit{Theme Relevance}, \textit{Concept Relevance}, \textit{Task Validity}, and \textit{Comprehensibility} on 3-point Likert scales (e.g., Relevant / Partially Relevant / Irrelevant). Supplementary dimensions included \textit{Difficulty} (3-point) and \textit{Interestingness} (5-point). After finishing the task ratings, we asked each evaluator to perform pairwise comparison in terms of novelty. Each pair comprised one task from {\cotmethod} and one task from {\dcmethod}. We randomly sampled 20 pairs of tasks. For each pair, the evaluator selected the more novel task and wrote a brief justification.

\section{Results}
\label{sec:evaluation_results}
\begin{figure}[!t]
    \Description{Two line charts. The left chart shows automated Vendi scores scaling up to 50 tasks, with CreativeDC outperforming baselines. The right chart shows the expert-annotated number of task clusters, again with CreativeDC producing more diverse clusters than the COT baseline.}
    \centering
    \includegraphics[width=0.78\linewidth]{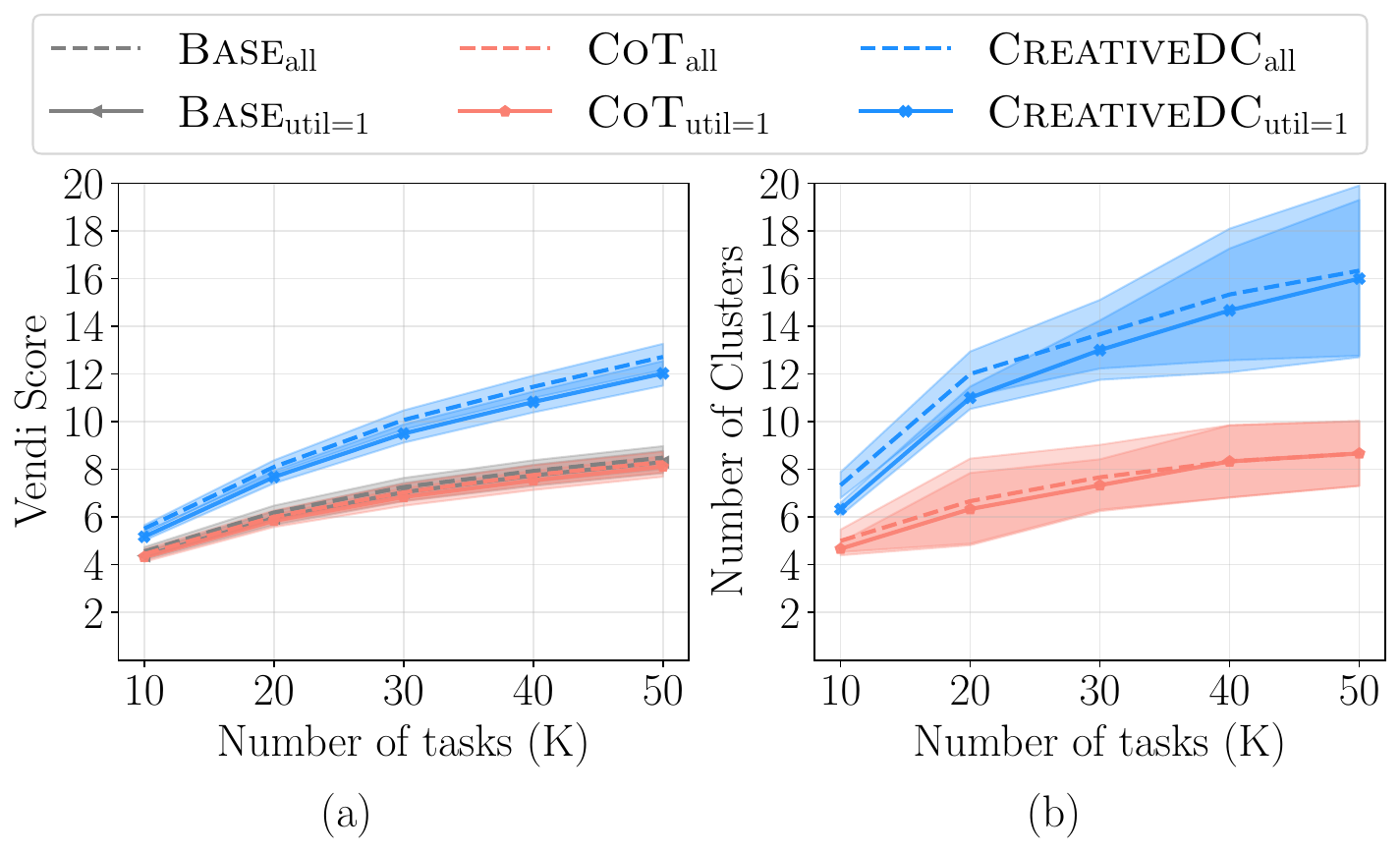}
    \caption{\textbf{Effective diversity with respect to number of tasks $K$} via (a) automated Vendi scores and (b) expert evaluation. Solid lines denote effective diversity (diversity of high-utility tasks only) and dashed lines denote diversity over all tasks. Shaded bands show SE across contexts. {\dcmethod} consistently outperforms baselines across both evaluations.}
    \label{fig:evaluation_vendiscore}
\end{figure}

We present the effective diversity with respect to the number of tasks $K$ in Figure~\ref{fig:evaluation_vendiscore} and ratings for other task aspects in Table~\ref{tab:expertstudyresults_1_2}.

\subsection{Effective Diversity}

\textbf{Effective Diversity via Automated Vendi Score.} Figure~\ref{fig:evaluation_vendiscore}a shows that all methods generate more distinct tasks as the number of tasks $K$ increases. Notably, {\dcmethod} significantly outperforms both {\cotmethod} and {\basemethod} across all $K$ values ($p < 0.001$, Bonferroni-corrected paired Wilcoxon signed-rank tests~\cite{c4091bd3-d888-3152-8886-c284bf66a93a}), while {\cotmethod} and {\basemethod} are statistically indistinguishable. At $K=50$, {\dcmethod} generates about $13$ distinct tasks (dashed blue line), and about $12$ distinct tasks when restricted to high-utility tasks only (solid blue line). In contrast, both {\cotmethod} and {\basemethod} generate about $8$ distinct high-utility tasks (solid red and grey lines).

\looseness-1\textbf{Effective Diversity via Expert Evaluation.} Figure~\ref{fig:evaluation_vendiscore}b shows that effective diversity perceived by experts has a similar pattern but is generally higher than the automated Vendi Score. For instance, evaluators created about $16$ clusters at $K=50$ for {\dcmethod} when restricted to high-utility tasks, compared to about $9$ clusters for {\cotmethod}. While {\cotmethod} shows signs of saturation, {\dcmethod} maintains an upward trend. Because each expert evaluated a uniquely themed subset of tasks, inter-rater agreement measures are not applicable. This evaluation provides evidence that the automated metric we used can serve as a proxy for expert-perceived diversity.

\subsection{Expert Ratings for Other Task Aspects}
We compute the average expert ratings (cf. Section~\ref{sec:measuring_other_task_aspects}) by mapping 3-point scales to $\{1, 0.5, 0\}$ and 5-point scales to $\{1, 0.75, 0.5, 0.25, 0\}$. Table~\ref{tab:expertstudyresults_1_2} shows that tasks generated by {\dcmethod} were judged higher than those from {\cotmethod} on the utility-related submetrics. Tasks generated by both methods were perceived as between easy and medium in difficulty. Notably, {\dcmethod} tasks were rated higher in interestingness ($0.81$ vs.\ $0.50$) and were preferred in $96.67\%$ of pairwise novelty comparisons. Justifications for experts' decisions favoring tasks generated by {\dcmethod} include: ``It makes the cooking theme and the question very life-like''; ``Trading recorded personal experiences as currency is definitely novel to me--the problem even opens up my mind for new possibilities that may really happen in the future''.

\begin{table}[!t]
    \centering
    \caption{Expert evaluator ratings (means $\pm$ SE) for various task aspects: utility, novelty, and difficulty. Highest scores are bolded for metrics where higher is better ($\uparrow$).}
    \label{tab:expertstudyresults_1_2}
    
        \centering
        
        \resizebox{0.70\linewidth}{!}{
        \renewcommand{\arraystretch}{1.2}
        \begin{tabular}{p{4.5cm}cc}
        \toprule
        \textbf{Metric} & {\cotmethod} & {\dcmethod} \\
        \midrule
        Utility:TaskValidity $\uparrow$ & \phantom{0}0.78 $\pm$ \phantom{0}0.15 & \textbf{\phantom{00}1.00 $\pm$ 0.00} \\
        Utility:ThemeRelevance $\uparrow$ & \phantom{0}0.94 $\pm$ \phantom{0}0.06 & \textbf{\phantom{00}1.00 $\pm$ 0.00} \\
        Utility:ConceptRelevance $\uparrow$ & \phantom{0}0.89 $\pm$ \phantom{0}0.07 & \textbf{\phantom{00}0.94 $\pm$ 0.06} \\
        Utility:Comprehensibility $\uparrow$ & \phantom{0}0.72 $\pm$ \phantom{0}0.15 & \textbf{\phantom{00}0.83 $\pm$ 0.08} \\
        \midrule
        Novelty:Interestingness $\uparrow$ & \phantom{0}0.50 $\pm$ \phantom{0}0.04 & \textbf{\phantom{00}0.81 $\pm$ 0.04} \\
        Novelty:Win Rate (\%) $\uparrow$ & \phantom{0}3.33 $\pm$ \phantom{0}2.32 & \textbf{\phantom{0}96.67 $\pm$ 2.32} \\
        \midrule
        Difficulty:Self-rated & \phantom{0}0.17 $\pm$ \phantom{0}0.08 & \phantom{00}0.17 $\pm$ 0.08 \\
        Difficulty:Pass Rate (\%) & 88.89 $\pm$ 11.11 & 100.00 $\pm$ 0.00 \\
        Difficulty:Time Taken (mins) & \phantom{0}3.84 $\pm$ \phantom{0}0.55 & \phantom{00}3.38 $\pm$ 0.61 \\

        \bottomrule
        \end{tabular}
        }
\end{table}

    \FloatBarrier
\section{Concluding Discussion}
\label{sec:conclusion}
We proposed {\dcmethod}, a method for generating diverse and high-utility tasks in programming education. Through automated and expert evaluations, {\dcmethod} has been shown to produce more distinct high-utility programming tasks, improving the effective diversity of generated tasks.

Next, we discuss limitations of our work and suggest future directions. First, our evaluation focused on introductory programming concepts; extending this framework to other educational domains, such as mathematics or science, would demonstrate its generalizability. Second, while our expert study provides initial validation, it is limited by its sample size and demographic diversity; it would be useful to conduct large-scale studies to evaluate the tasks from the perspective of students. Third, the two stages in our method can be iterated in a loop to improve the quality of the generated tasks; exploring this iterative process is a promising direction. Fourth, our experiments focus on introductory Python programming concepts; it would be useful to check whether this diversity holds in harder problems with advanced concepts and whether diversity across difficulty levels can be enforced while maintaining the effective validity and utility of the generated tasks. Finally, our evaluation focused on theme and concept as a task context; expanding the context to include learner-state contextualization (e.g., difficulty, student profiles) is a natural next step for generating diverse and personalized tasks.
    \section*{Acknowledgement}
Funded/Co-funded by the European Union (ERC, TOPS, 101039090). Views and opinions expressed are however those of the author(s) only and do not necessarily reflect those of the European Union or the European Research Council. Neither the European Union nor the granting authority can be held responsible for them.
    
    \bibliographystyle{unsrt}
    \bibliography{main}
    
\end{document}